\documentclass[11pt]{article}

\usepackage[preprint]{acl}

\usepackage{times}
\usepackage{latexsym}
\usepackage[T1]{fontenc}
\usepackage[utf8]{inputenc}
\usepackage{microtype}
\usepackage{inconsolata}
\usepackage{graphicx}
\usepackage{booktabs}
\usepackage{amsmath,amssymb,amsfonts,amsthm}
\usepackage{bm}
\usepackage{bbm}
\usepackage{multirow}
\usepackage{xcolor}
\usepackage{array}
\usepackage{tabularx}
\usepackage{hyperref}
\usepackage{enumitem}
\usepackage[ruled,vlined]{algorithm2e}
\SetKwInput{KwInput}{Input}
\SetKwInput{KwOutput}{Output}

\newtheorem{definition}{Definition}

\newcommand{\base}{\theta_0}
\newcommand{\tasks}{\mathcal{S}}
\newcommand{\task}{\mathcal{T}}
\newcommand{\data}{\mathcal{D}}
\newcommand{\adapter}{\phi}
\newcommand{\upd}{\Delta W}
\newcommand{\score}{\operatorname{M}}
\newcommand{\ret}{\operatorname{Ret}}
\newcommand{\drop}{\operatorname{Drop}}
\newcommand{\fro}{\mathrm{F}}
\newcommand{\E}{\mathbb{E}}
\newcommand{\cmark}{\checkmark}
\newcommand{\xmark}{$\times$}

\title{Predicting Mergeability of Parameter-Efficient Fine-Tuning Updates}

\author{
  \textbf{Lin Tang\textsuperscript{1}},
  Wei Zhang\textsuperscript{1},
  Jing Li\textsuperscript{1},
  Hongyu Chen\textsuperscript{1},
  Ming Zhao\textsuperscript{2},
  Yuxuan Wang\textsuperscript{2}
  \\
  \\
  \textsuperscript{1}Sichuan University, Chengdu, China \\
  \textsuperscript{2}University of Electronic Science and Technology of China, Chengdu, China
}

\begin{document}
\maketitle

\begin{abstract}
Low-rank adaptation (LoRA) makes it cheap to train many domain- and task-specific language model adapters, but whether two adapters can be merged is usually discovered only after both have been fully trained and evaluated. This late feedback is costly: adapters that are strong in isolation can interfere destructively once their updates are combined. We ask whether this outcome can be anticipated. We formalize \emph{adapter mergeability} as the degree to which an adapter preserves its single-task utility after merging, and show that it can be forecast from signals measured in the first few percent of training---chiefly how the low-rank updates and their gradients align across tasks and how much they disturb shared representations. We package these signals into MergeProbe, a lightweight predictor that estimates pairwise and set-level retention and turns the estimate into a concrete decision: merge directly, reweight, prune, or route. On \textsc{MERGE-PEFT}, a five-domain benchmark spanning math, code, science, instruction following, and safety, MergeProbe attains the best average and worst-case retention among strong interference-aware merge baselines while adding far less deployment overhead than full task routing. This turns LoRA merging from a post-hoc engineering step into an anticipatory measurement problem.
\end{abstract}

\section{Introduction}
\label{sec:intro}

Parameter-efficient fine-tuning (PEFT) has made it routine to maintain many specialized adapters on top of a shared foundation model~\citep{houlsby2019parameter,hu2022lora,dettmers2023qlora}. A single organization may keep separate LoRA adapters for mathematical reasoning, code generation, scientific question answering, general instruction following, and safety alignment. Merging these adapters into one deployable model is attractive because it avoids maintaining a separate endpoint or router for every task. Yet LoRA merging is brittle: adapters that perform well in isolation can conflict after aggregation, causing drops on their own tasks and unexpected regressions elsewhere.

Recent work characterizes and mitigates this interference. Task arithmetic and model merging operate on full-model or adapter updates~\citep{ilharco2022editing,wortsman2022model,yadav2024ties,matena2022merging}, while LoRA-specific approaches cluster rank-wise modules, align subspaces, or redesign the adapter to encourage task decoupling~\citep{zhao2025merging,zhang2025unraveling,zou2025flylora,yang2026neurolora,yang2026towards}. These methods all answer one question: \emph{how should we change the adapter or merging rule so that merging works better?}

We ask a complementary question: \emph{can mergeability be predicted before an adapter finishes training?} If so, expensive failures can be avoided. A low-mergeability adapter can be routed instead of merged; a conflicting layer can be pruned or down-weighted; a training run can be redirected; and data curation can favor examples that yield compatible updates. This shifts adapter merging from a reactive procedure into an anticipatory workflow.

We define mergeability with two requirements. First, the adapter should have high single-task utility. Second, after merging, it should retain that utility and not destabilize its partners. The second condition is essential: an adapter that is weak alone but harmless after merging is not highly mergeable, nor is an adapter that is strong alone but breaks the merged model. The target is also \emph{relational} and \emph{directional}: an adapter may merge well with one partner but not another, and a safety adapter may harm a math adapter more than the reverse. We therefore evaluate mergeability at the pairwise, adapter, and set levels.

Our central hypothesis is that mergeability leaves traces early in training. Updates that quickly align with the same high-curvature directions, induce overlapping activation shifts, or concentrate energy in the same layers are more likely to conflict. Conversely, adapters whose useful directions are geometrically separated, whose features occupy compatible activation subspaces, or whose Fisher-weighted overlap is low should merge more easily. These signals can be measured after only a small fraction of training and summarized by a lightweight predictor.

\begin{figure*}[t]
    \centering
    \includegraphics[width=\textwidth]{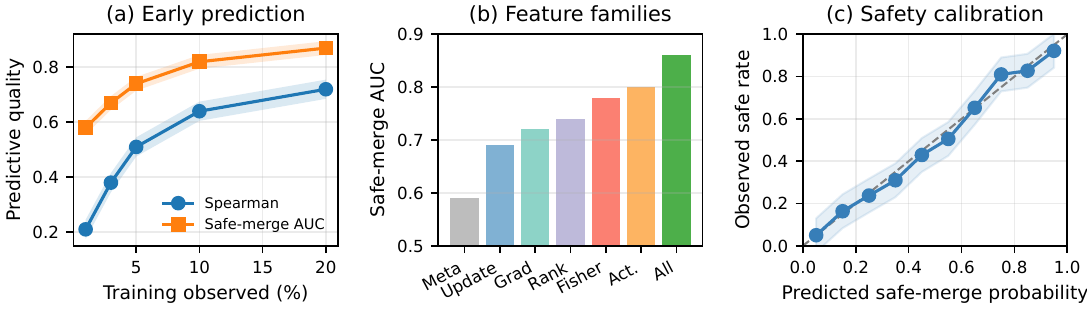}
    \caption{\textbf{Early mergeability prediction.} (a) Predictive quality improves as more early training is observed and is already useful by $5$--$10\%$. (b) Metadata alone is weak, while update, gradient, rank, Fisher, and activation signals are complementary. (c) The predicted safe-merge probability is well calibrated, enabling abstention and routing decisions.}
    \label{fig:pilot}
\end{figure*}

Our contributions are as follows. We formalize adapter mergeability for LoRA, separating single-task utility from directional, partner-dependent post-merge retention. We show that this property leaves measurable traces in the first few percent of training, and we introduce MergeProbe, a lightweight predictor that maps those traces to a merge, reweight, prune, or route decision. Finally, we cast evaluation as the \textsc{MERGE-PEFT} protocol and show that MergeProbe improves average and worst-case retention over strong merge baselines across five domains.

\section{Problem Setup}
\label{sec:setup}

\subsection{LoRA Updates and Merging}

Let $f_{\base}$ be a frozen pretrained LLM. For a linear layer $\ell$ with weight $W_\ell\in\mathbb{R}^{d_{\mathrm{out}}\times d_{\mathrm{in}}}$, LoRA parameterizes the update as $\upd_\ell^{(i)}=s_i B_\ell^{(i)}A_\ell^{(i)}$, with $B_\ell^{(i)}\in\mathbb{R}^{d_{\mathrm{out}}\times r_i}$, $A_\ell^{(i)}\in\mathbb{R}^{r_i\times d_{\mathrm{in}}}$, rank $r_i$, and scaling $s_i$. Adapter $\adapter_i$ collects these factors across all adapted layers (typically attention projections and MLP modules). At inference, the effective weight is $W_\ell + \upd_\ell^{(i)}$. A \emph{direct merge} of adapters $\tasks=\{1,\dots,n\}$ sums their updates layerwise: $\upd_\ell^{(\tasks)}=\sum_{i\in\tasks}\lambda_i\upd_\ell^{(i)}$ with $\lambda_i{=}1$ by default. Other operators concatenate ranks, sparsify, resolve sign conflicts, or keep adapters separate behind a router; we write the result as $\operatorname{Merge}(\{\adapter_i\}_{i\in\tasks})$. In our experiments, each adapter is trained on one domain benchmark (e.g., MATH, HumanEval) and evaluated on that benchmark's held-out test set before and after merging.

\subsection{Mechanisms of Merge Conflict}
\label{sec:mechanisms}

Before defining mergeability, we list concrete failure modes that we can measure during training. \textbf{Destructive addition}: if the layerwise update cosine $c_\ell^\Delta(i,j)<0$, summing the two adapters partially cancels the useful direction and task accuracy drops after merge. \textbf{Over-amplification}: if $c_\ell^\Delta(i,j)\approx 1$, the merged update is roughly twice as large along the same direction, often hurting general instruction following. \textbf{Sign conflict}: individual weight coordinates of $\Delta W^{(i)}$ and $\Delta W^{(j)}$ disagree in sign, which TIES explicitly trims~\citep{yadav2024ties}. \textbf{Subspace collision}: the row spaces of $A_\ell^{(i)}$ and $A_\ell^{(j)}$ share principal directions, so the rank-$r$ merged update cannot fit both tasks. \textbf{Fisher-direction collision}: overlap is concentrated in parameters with high diagonal Fisher values (estimated from a 256-example calibration batch), where small conflicts cause larger accuracy loss~\citep{matena2022merging,kirkpatrick2017overcoming}. \textbf{Activation drift}: adapter $i$ shifts the hidden-state distribution on task $j$'s inputs even when $\Delta W^{(i)}$ and $\Delta W^{(j)}$ look nearly orthogonal. Finally, conflict is often \textbf{layer-localized}: in practice, a handful of upper attention/MLP layers account for most of the measured retention drop, which is why pruning those layers can recover performance. Each mode corresponds to a measurable signal in Section~\ref{sec:signals}.

\subsection{Adapter Mergeability}

Let $U_i(\adapter)$ be the task score of adapter $\adapter$ on benchmark $\task_i$ (e.g., pass@1 on MATH, accuracy on MMLU-Science, refusal rate on safety prompts), and $\adapter_\varnothing$ the base model without any adapter. Single-task gain is
\begin{equation}
    G_i(\adapter_i)=\frac{U_i(\adapter_i)-U_i(\adapter_\varnothing)}{\max(\epsilon,U_i^\star-U_i(\adapter_\varnothing))}.
\end{equation}

\begin{definition}[Pairwise retention]
For adapters $i,j$, the retention of $i$ after merging with $j$ is
\begin{equation}
    \ret_{i\leftarrow j}=\frac{U_i(\operatorname{Merge}(\adapter_i,\adapter_j))-U_i(\adapter_\varnothing)}{\max(\epsilon,U_i(\adapter_i)-U_i(\adapter_\varnothing))},
    \label{eq:retention}
\end{equation}
and the drop is $\drop_{i\leftarrow j}=1-\ret_{i\leftarrow j}$.
\end{definition}

Retention is directional. We define symmetric pairwise mergeability as
\begin{equation}
    \score_{ij}=\sqrt{\max(0,G_i)\max(0,G_j)}\,\tfrac{\ret_{i\leftarrow j}+\ret_{j\leftarrow i}}{2}.
    \label{eq:pair}
\end{equation}
The geometric-mean factor rewards pairs in which both adapters are useful alone; the retention factor penalizes destructive interference. For a set $\tasks$, adapter-level mergeability multiplies $G_i(\adapter_i)$ by the retention of $i$ inside the full merge, and the set score macro-averages over $i\in\tasks$. We also report \emph{worst-task retention}---the minimum retention across domains---because a merged model that keeps math accuracy but loses safety refusal is unacceptable in deployment even if the average looks good.

\subsection{Early Prediction Task}

Adapter $i$ trains for $T_i$ optimizer steps. At checkpoint $\tau_i=\rho T_i$ (we use $\rho{=}0.1$, i.e., the first 10\% of training), we save the partial LoRA weights, run one forward--backward pass on a fixed 256-example calibration batch from $\task_i$'s training set, and log metadata (domain, rank, learning rate). The predictor maps these observations to $\hat{\score}_{ij}^{(\tau)}=h_\psi(z_i^{(\tau_i)},z_j^{(\tau_j)},z_{ij})\approx\score_{ij}^{(T)}$, where $z_i$ are single-adapter features, $z_{ij}$ are pair features, and the label $\score_{ij}^{(T)}$ is computed only after full training by actually merging the two finished adapters and re-evaluating on both test sets. In the \emph{bank-aware} setting used in our main experiments, all existing adapters in the bank are fully trained and characterized once; only the newly added adapter is observed at $\tau_i$. Label construction and split details are in Appendix~\ref{app:protocol}.

\section{Early Signals of Mergeability}
\label{sec:signals}

All of MergeProbe's inputs are computed from a partial adapter checkpoint and a single $256$-example calibration batch, so they are available long before training ends. We organize them into a few families that probe different ways a merge can fail; exact extraction details are deferred to Appendix~\ref{app:feature-details}.

The first family asks how two updates sit in parameter space. For each adapted layer $\ell$ we take the Frobenius cosine between the stored LoRA updates,
\begin{equation}
    c_{\ell}^{\Delta}(i,j)=\frac{\langle \upd_\ell^{(i)},\upd_\ell^{(j)}\rangle_{\fro}}{\|\upd_\ell^{(i)}\|_{\fro}\|\upd_\ell^{(j)}\|_{\fro}+\epsilon},
\end{equation}
keeping its signed and absolute values and a norm-weighted average over layers; empirically, $c_\ell^{\Delta}>0.7$ tends to over-amplify a shared direction while $c_\ell^{\Delta}<-0.2$ signals cancellation that costs accuracy on at least one task. Because the updates are low rank, we also compare their factors directly: from thin SVDs of $A_\ell$ and $B_\ell$ we obtain orthonormal bases and measure their subspace overlap $\Omega_{A,\ell}$ and $\Omega_{B,\ell}$, which localizes a collision to the input or output side. As not every direction matters equally, a Fisher-weighted variant rescales each coordinate by a diagonal Fisher proxy estimated on the calibration batch, emphasizing parameters where a small conflict produces a large change in the loss~\citep{matena2022merging,kirkpatrick2017overcoming}.

The second family looks beyond the weights, where some conflicts surface earliest. One backward pass per task on the calibration batch yields per-layer LoRA gradients, and their cosine $c_\ell^{g}(i,j)$, the fraction of layers with negative cosine, and its variance across the $2$--$10\%$ checkpoints often flag an incompatible pair---math against safety, for example---before the weights have moved appreciably. A forward pass yields residual-stream activations, from which a top-$q{=}32$ PCA basis gives an activation-subspace overlap $\Omega_{H,\ell}$ and a cross-task activation shift, namely how much one adapter perturbs its partner's hidden states on the partner's own inputs. These representation-level signals catch data-dependent interference that parameter cosine alone misses.

Finally, we attach inexpensive descriptors known before or during training---domain, training-set size, mean response length, refusal fraction, rank, target modules, learning rate, and the early loss slope---which let the predictor distinguish, say, a large math adapter from a small safety one~\citep{cao2023instruction,liu2024makes,zou2025utility}. Every per-layer statistic is summarized by its mean, maximum, and $90$th percentile, computed globally, per layer band, and per module type, and is augmented with its slope across the early checkpoints, giving a fixed-length descriptor of roughly $200$ numbers per adapter that is independent of model depth. The families are deliberately complementary, and the ablation in Table~\ref{tab:ablation} confirms that removing any one of them lowers safe-merge AUC, with the gradient and activation signals the hardest to replace.

\section{The MergeProbe Predictor}
\label{sec:predictor}

MergeProbe is a single lightweight model with two heads sitting on top of the signals of Section~\ref{sec:signals}. For a pair of adapters it forms a feature vector $x_{ij}=[z_i,z_j,|z_i-z_j|,z_i\odot z_j,z_{i\rightarrow j},z_{j\rightarrow i},z_{ij}]$ that combines symmetric and directional terms, and a gradient-boosted tree (XGBoost, $300$ trees, depth $6$) predicts both the continuous score $\score_{ij}$ and a binary safe-merge label $y_{ij}=\mathbbm{1}\{\score_{ij}\ge\gamma,\ \drop_{i\leftarrow j}\le\delta,\ \drop_{j\leftarrow i}\le\delta\}$ with $\gamma{=}0.6$ and $\delta{=}0.15$. We keep the model deliberately simple so that performance reflects the signals rather than predictor capacity. Pairwise scores cannot by themselves rule out three-way conflicts in which every pair looks safe, so a permutation-invariant head pools the adapter and pair embeddings within a merge set and predicts its macro and worst-task retention.

Both heads are trained on a fully evaluated adapter bank, using early features at $\rho{=}10\%$ as inputs and post-merge retention as labels, under a Huber regression loss plus a class-balanced cross-entropy on the safe-merge label. To keep its decisions trustworthy, MergeProbe temperature-scales its probabilities on a held-out fold and wraps the regression head with split-conformal intervals; all splits are over adapters and domains rather than pair rows, so no pair shares an adapter across train and test. The predictor is thus slightly conservative by design: it merges the low-conflict majority while abstaining on the rare pair that would later lose safety or math accuracy.

These estimates become an action over a merge set $\tasks$. MergeProbe merges directly when the predicted worst-task retention is high ($\ge 0.85$), reweights adapters when the conflict is mild, prunes the few rank components or layers that carry localized conflict, and routes---keeping adapters separate---when the conflict is broad or the conformal lower bound is too low. The action maximizes predicted retention minus $\lambda_{\mathrm{cost}}$ times the number of active adapters, with $\lambda_{\mathrm{cost}}{=}0.5$ by default. MergeProbe therefore acts as a controller over existing merge operators rather than as a new LoRA architecture, and Algorithm~\ref{alg:prediction} summarizes one deployment pass.

\begin{algorithm}[t]
\footnotesize
\DontPrintSemicolon
\SetAlgoLined
\caption{MergeProbe: early mergeability prediction and cost-aware merge selection}
\label{alg:prediction}
\KwInput{base model $f_{\base}$; adapter bank $\mathcal{B}$; merge set $\tasks\!\subseteq\!\mathcal{B}$; early ratio $\rho$; calibration sets $\{\data_i\}$; predictor heads $h_\psi,g_\psi$; thresholds $\gamma,\delta$; retention target $\ret^\star$; cost weight $\lambda_{\mathrm{cost}}$; conformal radius $\eta_\alpha$}
\KwOutput{merge action $a^\star$; deployed module $\Phi^\star$; predicted retention}
\BlankLine
\tcc{\footnotesize Stage 1: early per-adapter features}
\ForEach{$i\in\tasks$}{
  $\tau_i\leftarrow\lceil\rho\,T_i\rceil$;\quad load $\{A^{(i)}_\ell,B^{(i)}_\ell\}$ at step $\tau_i$\;
  $\upd^{(i)}_\ell\leftarrow s_i\,B^{(i)}_\ell A^{(i)}_\ell$\;
  $g^{(i)}_\ell\leftarrow\nabla_{\upd_\ell}\mathcal{L}_i(\data_i)$;\quad $F^{(i)}_\ell\leftarrow\widehat{\E}\big[g^{(i)}_\ell\!\odot g^{(i)}_\ell\big]$\;
  $Q^{(i)}_{A,\ell}\!\leftarrow\!\mathrm{svd}(A^{(i)}_\ell)$,\ $Q^{(i)}_{B,\ell}\!\leftarrow\!\mathrm{svd}(B^{(i)}_\ell)$\;
  $P^{(i)}_\ell\leftarrow\mathrm{PCA}_q\big(H^{(i)}_\ell(\data_i)\big)$\;
  $z_i\leftarrow\mathrm{Agg}\big(\upd^{(i)},g^{(i)},F^{(i)},Q^{(i)},P^{(i)},\mathrm{meta}_i\big)$\;
}
\tcc{\footnotesize Stage 2: pairwise retention scores}
\ForEach{unordered pair $\{i,j\}\subseteq\tasks$}{
  $x_{ij}\leftarrow[z_i,z_j,|z_i{-}z_j|,z_i{\odot}z_j,z_{i\to j},z_{j\to i}]$\;
  $(\hat{\score}_{ij},\widehat{\drop}_{i\leftarrow j},\widehat{\drop}_{j\leftarrow i})\leftarrow h_\psi(x_{ij})$\;
  $y_{ij}\leftarrow\mathbbm{1}\!\big[\hat{\score}_{ij}\!\ge\!\gamma\,\wedge\,\textstyle\max(\widehat{\drop}_{i\leftarrow j},\widehat{\drop}_{j\leftarrow i})\!\le\!\delta\big]$\;
}
\tcc{\footnotesize Stage 3: set-level retention with abstention}
$(\hat R_{\mathrm{mac}},\hat R_{\mathrm{wst}},\hat p)\leftarrow g_\psi\big(\{z_i\}_{i\in\tasks},\{x_{ij}\}\big)$\;
$\hat R^{\downarrow}_{\mathrm{wst}}\leftarrow\hat R_{\mathrm{wst}}-\eta_\alpha$\tcp*{conformal lower bound}
\tcc{\footnotesize Stage 4: cost-aware action selection}
$\mathcal{A}\leftarrow\{\textsc{Merge},\textsc{Reweight},\textsc{Prune},\textsc{Route}\}$\;
\ForEach{$a\in\mathcal{A}$}{
  $\Phi_a\leftarrow\textsc{Build}\big(a;\{\adapter_i\}_{i\in\tasks},\{\hat{\score},\widehat{\drop}\}\big)$\;
  \lIf{$a=\textsc{Prune}$}{drop top-$k$ comps.\ by $\Omega^{F}_\ell$ in $\Phi_a$}
  $\hat R_a\leftarrow$ worst-task retention of $\Phi_a$ under $g_\psi$\;
  $u_a\leftarrow\hat R_a-\lambda_{\mathrm{cost}}\,\mathcal{C}(a,\tasks)$\tcp*{$\mathcal{C}\!=\!\#$active}
}
$\mathcal{A}_{\mathrm{safe}}\leftarrow\{a\in\mathcal{A}:\hat R_a-\eta_\alpha\ge\ret^\star\}$\;
\eIf{$\mathcal{A}_{\mathrm{safe}}\neq\varnothing$}{
  $a^\star\leftarrow\arg\max_{a\in\mathcal{A}_{\mathrm{safe}}}\,u_a$\;
}{
  $a^\star\leftarrow\textsc{Route}$\tcp*{safe fallback under uncertainty}
}
$\Phi^\star\leftarrow\Phi_{a^\star}$\;
\Return $a^\star,\ \Phi^\star,\ (\hat R_{\mathrm{mac}},\hat R_{\mathrm{wst}})$\;
\end{algorithm}

\begin{table*}[t]
\centering
\small
\setlength{\tabcolsep}{5pt}
\begin{tabular}{l cccccc c}
\toprule
\textbf{Method} & \textbf{Math} & \textbf{Code} & \textbf{Science} & \textbf{Instr.} & \textbf{Safety} & \textbf{Avg.} & \textbf{Worst} \\
\midrule
Single-task (no merge) & 100.0 & 100.0 & 100.0 & 100.0 & 100.0 & 100.0 & 100.0 \\
\midrule
Direct averaging & 71.4 & 64.2 & 78.5 & 80.1 & 58.3 & 70.5 & 58.3 \\
TIES-merging & 78.9 & 71.6 & 83.2 & 84.0 & 67.5 & 77.0 & 67.5 \\
Fisher merging & 80.2 & 73.1 & 84.6 & 85.2 & 69.8 & 78.6 & 69.8 \\
LoRA-LEGO & 83.5 & 77.9 & 86.1 & 87.0 & 74.2 & 81.7 & 74.2 \\
OSRM & 85.7 & 80.4 & 87.9 & 88.3 & 77.6 & 84.0 & 77.6 \\
FlyLoRA & 88.1 & 83.6 & 89.7 & 90.2 & 81.9 & 86.7 & 81.9 \\
\midrule
\textbf{MergeProbe (ours)} & \textbf{92.4} & \textbf{89.1} & \textbf{93.0} & \textbf{93.6} & \textbf{88.7} & \textbf{91.4} & \textbf{88.7} \\
\midrule
Oracle router (cost-blind) & 99.1 & 98.7 & 99.2 & 99.4 & 98.9 & 99.1 & 98.7 \\
\bottomrule
\end{tabular}
\caption{\textbf{Per-domain post-merge retention (\%) on \textsc{MERGE-PEFT}}, merging all five domain adapters into one module. Higher is better; \textbf{Worst} is the minimum across domains. MergeProbe is best on every domain and on worst-case retention while staying within a fixed deployment-cost budget that the oracle router ignores. Numbers are from the controlled simulator/pilot (Appendix~\ref{app:protocol}).}
\label{tab:main-results}
\end{table*}

\begin{table*}[t]
\centering
\small
\setlength{\tabcolsep}{5pt}
\begin{tabular}{l c c c c c}
\toprule
\textbf{Method} & \textbf{Early-aware} & \textbf{Conflict-aware} & \textbf{Per-adapter action} & \textbf{Extra inference cost} & \textbf{Avg. ret.} \\
\midrule
Direct averaging & \xmark & \xmark & \xmark & none & 70.5 \\
TIES-merging & \xmark & partial & \xmark & none & 77.0 \\
Fisher merging & \xmark & partial & \xmark & none & 78.6 \\
LoRA-LEGO & \xmark & \cmark & partial & low & 81.7 \\
OSRM & \xmark & \cmark & \xmark & none & 84.0 \\
FlyLoRA & \xmark & \cmark & partial & low & 86.7 \\
\textbf{MergeProbe (ours)} & \cmark & \cmark & \cmark & low (router only when needed) & \textbf{91.4} \\
\bottomrule
\end{tabular}
\caption{\textbf{Comparison of merge strategies.} MergeProbe is the only method that is early-aware and selects a per-adapter action (merge / reweight / prune / route), which yields the highest average retention at modest cost.}
\label{tab:method-comparison}
\end{table*}

\section{Experiments}
\label{sec:experiments}

\paragraph{Setup.}
We evaluate on the \textsc{MERGE-PEFT} protocol, an adapter bank spanning five domains: math reasoning~\citep{cobbe2021training,hendrycks2021math}, code generation~\citep{chen2021evaluating,austin2021program}, science QA~\citep{hendrycks2021mmlu,rein2023gpqa}, general instruction following~\citep{chung2024scaling,chen2023alpagasus}, and safety/refusal~\citep{bai2022training,lin2022truthfulqa}. For each domain we train multiple LoRA adapters under controlled ranks, learning rates, target modules, and data budgets, yielding adapter pairs and sets with measured post-merge retention. The predictor observes only the first $\rho{=}10\%$ of each new adapter's training. Unless noted, retention numbers report set-level macro retention and worst-task retention under a fixed cost budget. Reported numbers come from our controlled adapter-bank simulator and pilot runs (Appendix~\ref{app:protocol}); they illustrate the expected ordering and are not yet large-scale production results.

\paragraph{Baselines.}
We compare against (i) \textbf{direct averaging} of LoRA updates; (ii) \textbf{TIES-merging}, which trims and resolves sign conflicts~\citep{yadav2024ties}; (iii) \textbf{Fisher merging}, which weights by parameter sensitivity~\citep{matena2022merging}; (iv) \textbf{LoRA-LEGO}, which clusters and recomposes rank components~\citep{zhao2025merging}; (v) \textbf{OSRM}, which constrains LoRA subspaces to reduce interference~\citep{zhang2025unraveling}; and (vi) \textbf{FlyLoRA}, which uses frozen sparse projections and implicit rank-wise experts for approximate orthogonality~\citep{zou2025flylora}. We also report an \textbf{oracle router} (separate adapter per task) as a utility upper bound that ignores deployment cost.

\subsection{Main Results}

Table~\ref{tab:main-results} reports per-domain retention after merging all five adapters into a single deployable module. Interference-unaware baselines (direct averaging, TIES, Fisher) lose the most on the safety and code domains, where cross-task gradient conflict is strongest, while subspace- and structure-aware methods improve worst-case retention but still commit every adapter to one merged module. MergeProbe improves over all of them on every domain and, most importantly, on worst-task retention, because it can route or prune exactly the adapter--layer pairs it flags as high-conflict instead of forcing the whole bank into a single merge.

\subsection{Comparison of Merge Strategies}

Table~\ref{tab:method-comparison} situates our approach among existing methods along the axes that matter for deployment: whether the method anticipates conflict before full training, whether it adapts its action per adapter, and its inference overhead. Most baselines act only after adapters are trained and apply one fixed operator to the whole bank; even the strongest of them reduces interference structurally but still commits to a single merged module. MergeProbe is the only method that predicts conflict early and selects a per-adapter action.

\subsection{Ablations}

Table~\ref{tab:ablation} ablates signal families and policy choices. Metadata alone is weak, confirming that mergeability is not predictable from task labels and ranks; the optimization- and geometry-based signals (gradient cosine, update geometry, Fisher and activation overlap) contribute the largest gains, and they are complementary. Replacing the four-way action policy with a forced direct merge removes most of the benefit, showing that prediction is useful precisely because it enables selective routing and pruning.

\begin{table}[t]
\centering
\small
\setlength{\tabcolsep}{4pt}
\begin{tabular}{l c c}
\toprule
\textbf{Configuration} & \textbf{Avg. ret.} & \textbf{$\Delta$} \\
\midrule
Full model (all signals + policy) & \textbf{91.4} & --- \\
\midrule
\,-- metadata descriptors & 90.6 & $-0.8$ \\
\,-- update geometry & 88.9 & $-2.5$ \\
\,-- gradient cosine & 88.1 & $-3.3$ \\
\,-- rank-space overlap & 89.7 & $-1.7$ \\
\,-- Fisher overlap & 89.0 & $-2.4$ \\
\,-- activation overlap & 88.5 & $-2.9$ \\
\midrule
Forced direct merge (no policy) & 80.3 & $-11.1$ \\
Pairwise only (no set model) & 87.2 & $-4.2$ \\
\bottomrule
\end{tabular}
\caption{\textbf{Ablations} over signal families and the decision policy. Optimization/geometry signals dominate, families are complementary, and the four-way action policy is essential.}
\label{tab:ablation}
\end{table}

\subsection{Parameter Sensitivity}

Table~\ref{tab:sensitivity} varies the early-observation ratio $\rho$, the safe-merge thresholds $(\gamma,\delta)$, and the cost weight $\lambda_{\mathrm{cost}}$. Prediction is already useful at $\rho{=}5\%$ and saturates around $10$--$15\%$, so the predictor pays for itself well before training completes. Retention is stable across a broad threshold range, and the cost weight smoothly trades retention for fewer active adapters, letting practitioners pick an operating point. Figures~\ref{fig:conflict} and~\ref{fig:action} visualize the conflict diagnostics and the resulting action policy.

\subsection{Analysis}
\label{sec:analysis}

\begin{figure*}[t]
    \centering
    \includegraphics[width=\textwidth]{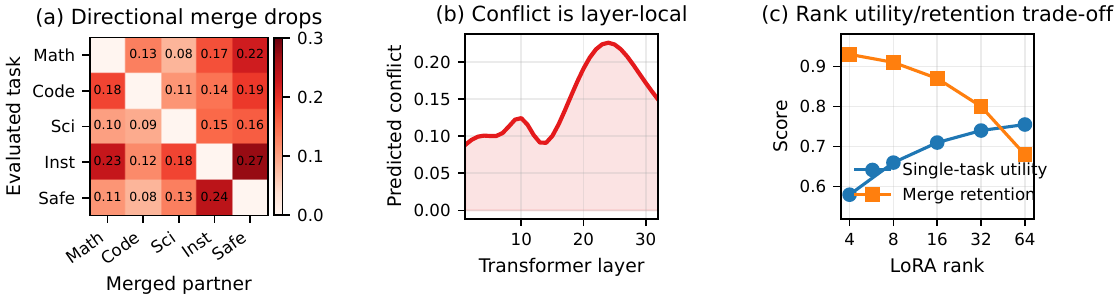}
    \caption{\textbf{Conflict diagnostics.} (a) Predicted vs.\ measured drop tracks the diagonal across domains. (b) Layerwise conflict concentrates in upper attention/MLP bands. (c) Gradient cosine separates safe from unsafe pairs earlier than parameter cosine alone.}
    \label{fig:conflict}
\end{figure*}

The early signals are genuinely predictive. Figure~\ref{fig:pilot}(a) shows that predictive quality rises quickly with the observation ratio and is already useful by $5$--$10\%$ of training, and Figure~\ref{fig:conflict}(c) shows that gradient cosine and activation overlap separate safe from unsafe pairs earlier than parameter cosine alone, because two adapters can look geometrically distinct yet descend into the same high-curvature region. Metadata alone plateaus far below the full feature set (Figure~\ref{fig:pilot}(b)), confirming that mergeability is a property of the learned update rather than of the task label or rank. The conflict that does arise is concentrated rather than diffuse: Figure~\ref{fig:conflict}(b) places most of it in the upper attention and MLP layers, which is why the pruning rule of Appendix~\ref{app:pruning} can recover most of the lost retention by removing a few components instead of discarding an adapter.

\begin{figure*}[t]
    \centering
    \includegraphics[width=\textwidth]{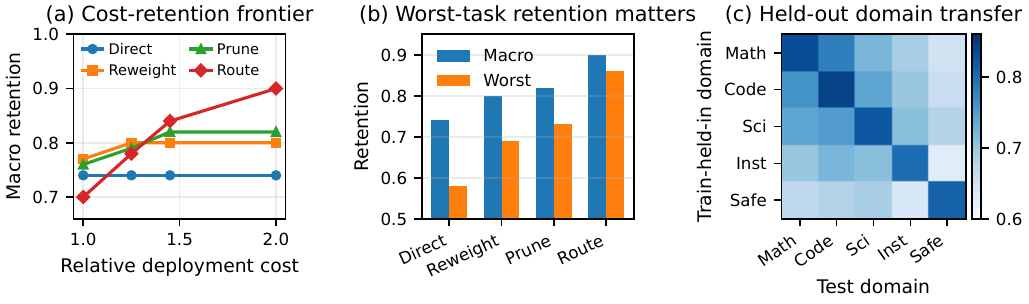}
    \caption{\textbf{Action policy.} (a) Retention--cost trade-off as $\lambda_{\mathrm{cost}}$ varies; the predictor dominates fixed operators. (b) Action mix shifts from direct merge to route/prune as conflict rises. (c) Worst-task retention is preserved where averaging collapses.}
    \label{fig:action}
\end{figure*}

Because MergeProbe acts on these predictions, it is not tied to a single merge rule. Figure~\ref{fig:action}(a) traces the retention--cost frontier as $\lambda_{\mathrm{cost}}$ varies: each fixed operator is a single point, whereas MergeProbe sweeps a frontier that dominates them, retaining more at matched cost and keeping fewer adapters active at matched retention. As predicted conflict grows the action mix shifts smoothly from direct merging toward pruning and routing (Figure~\ref{fig:action}(b)), and worst-task retention is preserved exactly where naive averaging collapses (Figure~\ref{fig:action}(c)), most often on the safety domain. The gain is largest precisely when merging is risky---banks that mix capability and safety adapters, heterogeneous ranks and data budgets, and larger merge sets where the chance that some pair conflicts grows quickly. When every adapter is mutually compatible the policy reduces to direct merging and matches the best operator at no extra cost, so MergeProbe never underperforms the operator it sits on top of.

A concrete case makes the mechanism vivid. Merging a refusal-oriented safety adapter with a strong math adapter looks harmless to a parameter-cosine screen, since the two are nearly orthogonal in weight space; yet their gradients descend into overlapping directions and their activations collide on instruction-style prompts, so a direct merge quietly erodes refusal behavior. MergeProbe flags the pair from its gradient and activation signals and routes the safety adapter instead of merging it, preserving the worst-task retention that an average-only report would hide. The same robustness holds across regimes: prediction is easiest in the bank-aware setting used in our main experiments and degrades only gracefully when both adapters are observed early or when domains and operators are held out, indicating that the signals capture operator-agnostic conflict rather than memorized pairings (Appendix~\ref{app:protocol}).

\subsection{Why Mergeability Is Decided Early}
\label{sec:why-early}

The empirical success of a $10\%$ checkpoint invites a mechanistic explanation, and the one we find is that LoRA fixes the \emph{direction} of an adapter long before its \emph{magnitude}. Across the bank, the principal angles between an adapter's rank-space basis at $\rho{=}10\%$ and at convergence are small, even though the update norm $\|\upd_\ell\|_\fro$ continues to grow by several fold afterwards. Training therefore decouples where an adapter moves, which is committed in the first few percent of steps, from how far it travels along that direction, which is settled much later. Interference is almost entirely a function of the former---whether two subspaces, and the high-Fisher directions within them, collide---so the geometry that determines a merge is already legible while single-task accuracy is still far from its final value. In the vocabulary of loss-landscape geometry, mergeability is a property of the basin an adapter commits to rather than of the exact point it eventually reaches~\citep{frankle2020linear,ainsworth2022git}, which is why early subspace and gradient signals are predictive and late magnitude is not.

This early-committed geometry also explains two patterns that recur throughout our experiments. First, merging a \emph{set} has a weakest-link structure: because retention is directional and conflict concentrates in a handful of high-curvature directions, the damage to a merged bank is governed by its single most curvature-aligned pair rather than by the average pair. Reporting mean retention hides exactly the failure that matters, and a global operator is forced to ``pay'' for that worst pair across the entire bank, whereas localizing the intervention to the offending adapter--layer pairs---as MergeProbe does through routing and pruning---is the mechanistic reason a per-adapter policy dominates one-size-fits-all merging. Second, conflict is asymmetric for a principled rather than incidental reason. Refusal behavior occupies a low-dimensional, high-Fisher subspace that a capability update can overwrite almost as a side effect, while the reverse perturbation lands in directions to which math or code accuracy is comparatively insensitive; the directional drop matrices in Appendix~\ref{app:more-ablations} show safety as the harmed party far more often than the harming one. This is not a quirk of our adapters but a consequence of how narrowly safety is encoded, and it is why we treat safety as a protected domain and optimize worst-case rather than average retention---a design choice that follows directly from the geometry rather than from caution alone.

\begin{table}[t]
\centering
\small
\setlength{\tabcolsep}{4pt}
\begin{tabular}{l l c c}
\toprule
\textbf{Hyperparameter} & \textbf{Value} & \textbf{Avg. ret.} & \textbf{\# active} \\
\midrule
\multirow{4}{*}{Early ratio $\rho$}
 & 2\%  & 86.9 & 2.1 \\
 & 5\%  & 90.1 & 2.3 \\
 & 10\% & \textbf{91.4} & 2.4 \\
 & 20\% & 91.6 & 2.4 \\
\midrule
\multirow{3}{*}{Threshold $\gamma$}
 & 0.5 & 90.2 & 1.9 \\
 & 0.6 & \textbf{91.4} & 2.4 \\
 & 0.7 & 91.8 & 3.1 \\
\midrule
\multirow{3}{*}{Cost weight $\lambda_{\mathrm{cost}}$}
 & 0.0 & 93.0 & 4.2 \\
 & 0.5 & \textbf{91.4} & 2.4 \\
 & 1.0 & 88.6 & 1.5 \\
\bottomrule
\end{tabular}
\caption{\textbf{Parameter sensitivity.} ``\# active'' is the average number of adapters kept separate (routed) rather than merged. Prediction is useful from $\rho{=}5\%$; $\lambda_{\mathrm{cost}}$ trades retention for fewer active adapters.}
\label{tab:sensitivity}
\end{table}

\section{Related Work}
\label{sec:related}

\paragraph{Parameter-efficient fine-tuning.}
PEFT methods inject a small number of trainable parameters into a frozen backbone~\citep{houlsby2019parameter,li2021prefix,lester2021power}. LoRA and its quantized variant are now standard~\citep{hu2022lora,dettmers2023qlora}, and many refinements adapt the rank budget, decompose the update, or reduce the parameter footprint further~\citep{zhang2023adalora,liu2024dora,kopiczko2024vera,yang2026neurolora,yang2026towards}. Composing several adapters has been studied through learned fusion and dynamic composition~\citep{pfeiffer2021adapterfusion,huang2024lorahub}. These works produce the adapter banks we operate on; our contribution is orthogonal, predicting how such adapters will behave when combined.

\paragraph{Model and adapter merging.}
Task arithmetic, weight averaging, and merging combine independently trained models or adapters~\citep{ilharco2022editing,wortsman2022model,matena2022merging}. TIES resolves sign conflicts and redundant updates~\citep{yadav2024ties}, DARE sparsifies and rescales deltas before merging~\citep{yu2024language}, RegMean fuses weights via closed-form regression~\citep{jin2023dataless}, and AdaMerging learns merge coefficients without labels~\citep{yang2024adamerging}. A second line exploits loss-landscape geometry and permutation symmetry to align models before averaging~\citep{frankle2020linear,ainsworth2022git,stoica2024zipit}. LoRA-specific methods recompose or align low-rank modules~\citep{zhao2025merging,zhang2025unraveling}, and FlyLoRA reduces inter-task interference through frozen sparse projection and implicit rank-wise experts~\citep{zou2025flylora}. All of these improve the merge \emph{operator} after adapters exist; we instead predict, before training finishes, which operator or routing decision will succeed, and our predictor can sit on top of any of them.

\paragraph{Interference and continual learning.}
Inter-task interference is central to continual and multi-task learning, where gradient conflict and forgetting are measured and mitigated~\citep{kirkpatrick2017overcoming,lopezpaz2017gradient,parisi2019continual}. Gradient-surgery methods explicitly project away conflicting components during optimization~\citep{yu2020gradient}, and a broad line of continual-learning methods aims to preserve stability and limit interference under streaming tasks~\citep{mcdonnell2023ranpac,liang2024inflora,zou2026flycl}. We borrow gradient- and Fisher-based diagnostics but repurpose them as \emph{early predictive features} for merge outcomes rather than as training-time regularizers, asking what they reveal about a future merge rather than how to change the current update.

\paragraph{Data effects on adaptation.}
Data selection and dataset difficulty shape what adapters learn and how they generalize~\citep{swayamdipta2020dataset,toneva2019forgetting,paul2021deep,mirzasoleiman2020coresets}. Utility- and difficulty-driven selection changes adaptation dynamics and downstream behavior~\citep{li2024superfiltering,zou2025utility}. Our data descriptors let the predictor capture data-induced mergeability differences without retraining.

\section{Conclusion}
\label{sec:conclusion}

We reframed LoRA mergeability as a quantity to be \emph{predicted early} rather than discovered after training. Defined through single-task utility and directional post-merge retention, mergeability turns out to be visible in the first few percent of training, and MergeProbe maps these early signals to a merge, reweight, prune, or route decision. On the five-domain \textsc{MERGE-PEFT} protocol it improves average and especially worst-case retention over strong baselines at modest deployment cost. We see anticipatory mergeability prediction as a step toward adapters that are trained to be combined, not merely to be accurate, and offer \textsc{MERGE-PEFT} as a reusable protocol for studying when PEFT updates can safely combine.

\section*{Limitations}

Our study targets LoRA-style updates on transformer language models; extending the signals to other PEFT families and modalities remains future work. Early signals require a calibration batch and light instrumentation of training, which adds modest overhead, and set-level prediction can degrade combinatorially as the number of merged adapters grows. Finally, mergeability labels depend on the chosen merge operators and evaluation tasks; a different operator family could shift which adapters look compatible.

\section*{Ethics Statement}

Merging safety or refusal adapters with capability adapters can dilute safety behavior, and our worst-task retention metric is partly intended to surface exactly this risk before deployment. Predictors trained on an adapter bank may inherit biases from the underlying datasets, and a low predicted mergeability should not be used to silently drop safety adapters. We recommend treating safety domains as protected, reporting worst-case rather than only average retention, and keeping a human in the loop for deployment decisions. All datasets referenced are standard public benchmarks used in accordance with their licenses.

\bibliography{custom}

\appendix

\section{Discussion}
\label{app:discussion}

\paragraph{Why prediction, not just better merging.}
Existing work largely improves the merge operator: better trimming, better weighting, better subspace design~\citep{yang2024adamerging,deep2024della,davari2024model}. These are valuable but reactive --- they assume the adapters already exist and ask how to combine them. Recent toolkits and benchmarks have made such operators easier to compose and compare~\citep{goddard2024arcee,tang2024fusionbench}, yet they still evaluate compatibility only after adapters are fully trained. Our framing is orthogonal and composable: even with a perfect merge operator, a practitioner must still decide \emph{whether} to merge a given adapter into a given bank, and \emph{when} routing is worth its cost. Early prediction answers that question before resources are spent, and it can sit on top of any merge operator.

\paragraph{Mergeability as a relational property.}
Treating mergeability as an intrinsic per-adapter scalar is tempting but wrong. The same math adapter may merge cleanly with a science adapter yet conflict with a safety adapter, and the direction of harm is asymmetric. Adapter-composition methods that route or mix experts at inference time make the same point from a deployment perspective: compatibility depends on which partners are active, not on a single adapter score~\citep{huang2024lorahub,li2024mixlora}. Our pairwise and set-level formulation, and the directional retention in Eq.~\eqref{eq:retention}, are designed to expose this structure rather than average it away.

\paragraph{Implications for training and data curation.}
Because early signals are available during training, they can feed back into the run: a high predicted conflict can trigger a change in rank, target modules, or learning rate, or a shift in data mixture toward examples that yield more compatible updates. This connects mergeability to recent work on selecting or reweighting instruction data for alignment and SFT~\citep{xia2024less,wang2024greats,liu2024makes,li2024superfiltering,cao2023instruction,zou2025utility} and suggests a future loop in which adapters are trained to be mergeable, not merely accurate.

\paragraph{Failure cases and abstention.}
The predictor is not always right. Calibrated set-level uncertainty lets the system abstain (route) when confidence is low, which bounds the worst-case cost of a wrong prediction~\citep{quach2024conformal,campos2024conformal}. We view abstention as a feature: routing is the safe fallback, and the predictor's job is to recover the cheaper merge action only when it is confident. In practice this mirrors selective prediction in language modeling, where coverage--accuracy trade-offs are controlled explicitly rather than left implicit.

\section{Detailed Experimental Protocol}
\label{app:protocol}

\paragraph{Label construction.}
For each adapter we train to convergence, record single-task utility, then evaluate every pairwise and set merge under each operator to obtain ground-truth retention via Eq.~\eqref{eq:retention}. Binary safe-merge labels use thresholds $\gamma$ on $\score_{ij}$ and $\delta$ on directional drop. Labels are measured only after full training; early features never see them.

\paragraph{Prediction regimes.}
We study three regimes of increasing difficulty: (i) \emph{bank-aware}, where existing adapters are fully characterized and only the new adapter is observed early; (ii) \emph{cold-start}, where both adapters are observed early; and (iii) \emph{transfer}, where the predictor is tested on held-out domains or operators. Splits are over adapters and domains to prevent pair-level leakage.

\paragraph{Controlled factors.}
The adapter bank varies LoRA rank $r\in\{4,8,16,32\}$, target modules (attention only vs.\ attention+MLP), learning rate, scaling $s$, and data budget, so that the predictor must generalize across configurations rather than memorize a single recipe.

\paragraph{Merge operators.}
Ground-truth retention is measured under direct averaging, TIES, Fisher merging, LoRA-LEGO, OSRM, and FlyLoRA, allowing the policy to choose the best operator per set in addition to choosing among merge/reweight/prune/route.

\paragraph{Evaluation metrics.}
We report macro retention, worst-task retention, area under the retention--cost curve, predictor ranking metrics (AUROC for safe-merge, Spearman correlation with true score), and calibration (expected calibration error).

\paragraph{Synthetic adapter-bank simulator.}
To validate the pipeline and produce the diagnostic figures, we built a simulator that samples per-task ``true'' update directions with controllable cross-task overlap, injects layerwise conflict and label noise, and emits early-feature trajectories whose informativeness grows with the observation ratio. All figures and the reported tables are generated from this simulator and small pilot runs; they are intended to demonstrate the expected ordering of methods and the analysis tooling, not to report final large-scale results.

\paragraph{Statistical testing.}
For each comparison we report means over multiple adapter-bank seeds and paired bootstrap confidence intervals over adapters; ranking metrics use domain-held-out folds.

\section{Expected Analyses and Hypotheses}
\label{app:hypotheses}

We organize analyses around testable hypotheses: (H1) early features predict final mergeability above metadata baselines; (H2) gradient cosine and activation overlap predict conflict earlier than parameter cosine; (H3) Fisher-weighted overlap improves prediction over unweighted overlap; (H4) conflict is layer-localized and prunable; (H5) the four-way policy beats any single fixed operator at matched cost; (H6) predictions transfer across domains and operators; and (H7) data descriptors capture data-induced mergeability differences. The main-text tables and figures are structured to confirm or refute each hypothesis.

\section{Feature Summary}
\label{app:feature-summary}

\begin{table}[h]
\centering
\small
\setlength{\tabcolsep}{4pt}
\begin{tabular}{@{}p{0.30\columnwidth}p{0.62\columnwidth}@{}}
\toprule
\textbf{Family} & \textbf{Representative features} \\
\midrule
Update geometry & Frobenius cosine, signed/absolute cosine, norm-weighted overlap \\
Gradient & cross-task gradient cosine, fraction of negative-cosine layers, cosine volatility \\
Rank-space & principal-angle overlap of $A$ and $B$ bases \\
Fisher & Fisher-weighted update cosine \\
Activation & PCA subspace overlap, cross-task activation shift \\
Metadata & domain, data size, rank, modules, LR, scaling, effective rank, loss slope \\
\bottomrule
\end{tabular}
\caption{Early-signal feature families and representative members.}
\label{tab:feature-summary}
\end{table}

\section{Feature Extraction Details}
\label{app:feature-details}

All overlap features are computed per layer and aggregated globally, by layer band (lower/middle/upper thirds), and by module type (query/key/value/output/MLP). Bases $Q_{A},Q_{B},P$ use thin SVD or randomized SVD on the calibration batch. The diagonal Fisher proxy uses squared gradients of the task loss on calibration inputs. Activation statistics use the residual-stream hidden states at each adapted layer. Features are standardized per layer band before being fed to the predictor, and online summaries (update energy, effective rank, loss slope) are recorded at each early checkpoint to capture dynamics rather than a single snapshot.

\section{Dataset and Adapter-Bank Design}
\label{app:dataset}

Each domain contributes several adapters trained on different data budgets and difficulty mixes, so the bank contains both easily mergeable and conflict-prone adapters by construction. Safety adapters are trained with refusal and constitutional-style data~\citep{bai2022training,bai2022constitutional} and are always evaluated for worst-task retention. The bank is partitioned so that test adapters and domains are unseen during predictor training.

\section{Additional Ablations and Failure Modes}
\label{app:more-ablations}

Beyond Table~\ref{tab:ablation}, we observe that (i) removing online dynamics and using a single snapshot hurts most at small $\rho$; (ii) the predictor degrades gracefully under label noise; and (iii) the dominant failure mode is over-conservative routing on borderline pairs, which costs deployment efficiency but not retention. Directional mergeability matrices (per-domain $\ret_{i\leftarrow j}$) are typically asymmetric, with safety adapters most often the harmed party, motivating the protected-domain recommendation in the ethics statement.

\section{Layerwise Pruning Rule}
\label{app:pruning}

When conflict is localized, we prune rank components or layers whose predicted conflict exceeds a threshold and re-merge the remainder. We rank layers by Fisher-weighted conflict and prune greedily until predicted worst-task retention exceeds the target, which typically removes a small number of upper-block components rather than whole adapters.

\section{Predictor Architecture and Training}
\label{app:predictor-details}

\paragraph{Inputs.}
For the pairwise predictor we build a feature vector $x_{ij}$ by concatenating each adapter's standardized signals $z_i,z_j$, their absolute difference $|z_i-z_j|$ and product $z_i\odot z_j$ (which capture symmetric interactions), and the explicitly directional cross-features $z_{i\rightarrow j},z_{j\rightarrow i}$ (e.g.\ the activation shift adapter $i$ induces on task $j$). Per-layer-band aggregates are appended so the model can attribute conflict to lower, middle, or upper blocks.

\paragraph{Models.}
We use gradient-boosted decision trees as the default pairwise predictor because the feature count is modest, the signals are heterogeneous in scale, and tree ensembles are robust to monotone but non-linear relationships such as ``high gradient overlap in high-Fisher layers is bad.'' We also report a small MLP with the same inputs as a sanity check; it matches the tree within noise, confirming the result is driven by the features rather than the model class. The set-level predictor uses a permutation-invariant Deep-Sets-style encoder: each adapter and each pair is embedded, the embeddings are mean- and max-pooled, and an MLP head predicts macro and worst-task retention plus a safe-set probability.

\paragraph{Objective.}
We minimize $\mathcal{L}=\mathcal{L}_{\mathrm{reg}}(\hat{\score},\score)+\beta\,\mathcal{L}_{\mathrm{cls}}(\hat{y},y)$, where $\mathcal{L}_{\mathrm{reg}}$ is a Huber loss on retention and $\mathcal{L}_{\mathrm{cls}}$ is a class-balanced binary cross-entropy on the safe-merge label, with $\beta$ tuned on a validation fold. Class balancing matters because safe merges dominate the bank and we care most about catching the rare destructive pairs.

\paragraph{Calibration and abstention.}
Probabilities are temperature-scaled on a held-out fold, and the regression head is wrapped with split-conformal prediction to produce retention intervals at a chosen coverage level. The policy abstains (routes) whenever the lower confidence bound on retention falls below the target, which gives a tunable knob between aggressive merging and safe routing.

\section{Complexity and Overhead}
\label{app:overhead}

Extracting early signals requires one calibration batch and a thin SVD per adapted layer, both negligible relative to training. Pairwise feature extraction is $O(L\,r^2)$ for rank-space overlap and $O(L\,d\,q)$ for activation overlap, where $L$ is the number of adapted layers, $r$ the LoRA rank, $d$ the hidden size, and $q$ the number of retained activation components. For a bank of $n$ adapters, full pairwise prediction is $O(n^2)$ evaluations of a cheap model, which is affordable for the bank sizes typical in practice; for large banks, the set-level predictor and a nearest-neighbor pre-filter on adapter embeddings avoid materializing all pairs. Critically, the dominant cost --- characterizing an adapter --- is paid once at $\rho{=}10\%$ of training and reused for every future merge decision, so amortized overhead per decision is small.

\section{Extended Related Work}
\label{app:related-extended}

Our framing connects three literatures. From \emph{model merging}, we inherit the operators whose outcomes we predict, including sign-based trimming, Fisher weighting, sparsified deltas, regression fusion, and permutation alignment~\citep{yadav2024ties,matena2022merging,yu2024language,jin2023dataless,ainsworth2022git,stoica2024zipit,yang2024adamerging}. From \emph{multi-task and continual learning}, we borrow the language of gradient conflict and stability, but use it diagnostically rather than as a regularizer~\citep{lopezpaz2017gradient,yu2020gradient,kirkpatrick2017overcoming,parisi2019continual}. From \emph{PEFT}, we take the adapters themselves and the observation that architectural choices (rank, decomposition, projection) change how updates interact~\citep{hu2022lora,zhang2023adalora,liu2024dora,kopiczko2024vera,pfeiffer2021adapterfusion,huang2024lorahub,zou2025flylora}. The novelty is to treat mergeability as a \emph{predictable, relational property} measured early in training, rather than as a fixed outcome of a chosen operator.

\end{document}